\documentclass[10pt, svgnames]{article} 
\usepackage[preprint]{tmlr}

\usepackage{amsmath,amsfonts,bm}

\def\eqref#1{equation~\ref{#1}}

\def\1{\bm{1}}

\def\eps{{\epsilon}}

\DeclareMathAlphabet{\mathsfit}{\encodingdefault}{\sfdefault}{m}{sl}
\SetMathAlphabet{\mathsfit}{bold}{\encodingdefault}{\sfdefault}{bx}{n}

\usepackage{hyperref}
\usepackage{url}

\usepackage{xcolor}
\usepackage{tikz}
\usetikzlibrary{calc,math}
\usepackage{graphicx}
\usepackage{cleveref}
\usepackage{hyperref}
\usepackage[framemethod=TikZ]{mdframed}
\usepackage{mathtools, amsthm}
\usepackage{comment}
\usepackage{wrapfig}
\usepackage{tabularx}
\usepackage{multicol}
\usepackage{booktabs}
\usepackage{pifont}
\usepackage{adjustbox}
\usepackage[numbers, sort&compress]{natbib}

\newtheorem{definition}{Definition}

\newcommand{\cmark}{\ding{51}}
\newcommand{\xmark}{\ding{55}}
\newcommand{\dott}{\scriptsize\ding{108}}

\DeclareMathOperator{\pa}{\text{pa}}

\newcommand\independent{\protect\mathpalette{\protect\independenT}{\perp}}
\def\independenT#1#2{\mathrel{\rlap{$#1#2$}\mkern2mu{#1#2}}}

\pgfdeclarelayer{back}
\pgfdeclarelayer{front}
\pgfsetlayers{back,main,front}

\makeatletter
\pgfkeys{%
  /tikz/on layer/.code={
    \pgfonlayer{#1}\begingroup
    \aftergroup\endpgfonlayer
    \aftergroup\endgroup
  },
  /tikz/node on layer/.code={
    \gdef\node@@on@layer{%
      \setbox\tikz@tempbox=\hbox\bgroup\pgfonlayer{#1}\unhbox\tikz@tempbox\endpgfonlayer\egroup}
    \aftergroup\node@on@layer
  },
  /tikz/end node on layer/.code={
    \endpgfonlayer\endgroup\endgroup
  }
}

\def\node@on@layer{\aftergroup\node@@on@layer}

\makeatother

\hypersetup{%
    colorlinks = true,
    citecolor  = blue,
    linkcolor  = blue,
}

\newmdenv[ 
  linecolor=purple,
  linewidth=2pt,
  topline=false,
  bottomline=false,
  rightline=false,
  skipabove=\topsep,
  skipbelow=\topsep,
  backgroundcolor=purple!3
]{leftrule}

\newcounter{examplecounter}
\newenvironment{example}[1]
    {
    \refstepcounter{examplecounter}
    \vspace{15pt}
    \begin{leftrule}[
        frametitle={\textcolor{purple}{Example \theexamplecounter:} {\em #1}}
    ]}
    {
    \end{leftrule}
    \vspace{8pt}
    }

\crefname{examplecounter}{ex.}{exs.}
\Crefname{examplecounter}{Example}{Examples}

\title{Navigating causal deep learning\\\parbox{\textwidth}{\hfill\Large{\em a living paper}}}

\author{\name Jeroen Berrevoets \email jeroen.berrevoets@maths.cam.ac.uk \\
      \addr DAMTP, University of Cambridge
      \AND
      \name Krzysztof Kacprzyk \email kk751@cam.ac.uk \\
      \addr DAMTP, University of Cambridge
      \AND
      \name Zhoazhi Qian \email zhaozhi.qian@maths.cam.ac.uk\\
      \addr DAMTP, University of Cambridge
      \AND
      \name Mihaela van der Schaar \email mv472@cam.ac.uk\\
      \addr DAMTP, University of Cambridge\\
      The Alan Turing Institute}

\def\month{MM}  
\def\year{YYYY} 
\def\openreview{\url{https://openreview.net/forum?id=XXXX}} 

\begin{document}

%

\maketitle

\begin{abstract}
Causal deep learning (CDL) is a new and important research area in the larger field of machine learning. With CDL, researchers aim to structure and encode causal knowledge in the extremely flexible representation space of deep learning models. Doing so will lead to more informed, robust, and general predictions and inference--- which is important! However, CDL is still in its infancy. For example, it is not clear how we ought to compare different methods as they are so different in their output, the way they encode causal knowledge, or even how they represent this knowledge. 
This is a {\it living paper} that categorises methods in causal deep learning beyond Pearl's ladder of causation. We refine the rungs in Pearl's ladder, while also adding a separate dimension that categorises the parametric assumptions of both input and representation, arriving at \textbf{\em the map of causal deep learning}. Our map covers machine learning disciplines such as supervised learning, reinforcement learning, generative modeling and beyond.
Our paradigm is a tool which helps researchers to: find benchmarks, compare methods, and most importantly: identify research gaps. With this work we aim to structure the avalanche of papers being published on causal deep learning. While papers on the topic are being published daily, our map remains fixed. We open-source our map for others to use as they see fit: perhaps to offer guidance in a related works section, or to better highlight the contribution of their paper. 

\end{abstract}


\section{Introduction}

For a long time, philosophers and scientists have been formalising, identifying and quantifying causality in nature, even dating back to 18\textsuperscript{th} century philosopher David Hume \citep{hume1751philosophical}. The ability to perform causal reasoning has been recognized as a hallmark of human intelligence \citep{pearl2018book}. It is only expected that causality has found its way as a research topic in artificial intelligence (AI). In fact, we recognise a true explosion of ideas and proposals in machine learning (a sub-field of AI) that build upon ideas from causality.

{\bf Causality.} Although there exist several different frameworks for causal reasoning as well as some on-going philosophical debate \citep{lewis1974causation,lewis1979counterfactual,pearl2009causality,rebane2013recovery,simon1966cause,collingwood2001essay,gasking1955causation,menzies1993causation,von2004explanation,woodward2005making,salmon1984scientific,williamson2011mechanistic}, the most widely adopted notion of causality in computer science is given by the structural causal model (SCM), introduced in the early 20\textsuperscript{th} century \citep{wright1920relative} and now championed by Judea Pearl. These SCMs are the main topic of study in causality. They are used for identification, problem formulation, and reasoning. 

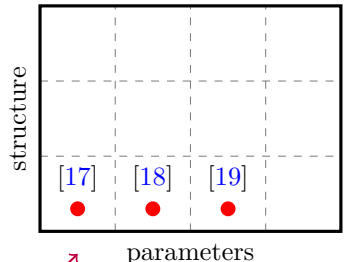
\begin{figure}[t]
    \centering
    \begin{tikzpicture}[
        remember picture,
        example/.style={circle, fill=red, inner sep=.7mm},
        grid/.style={gray, dashed},
    ]
        
        \node[inner sep=0, anchor=north west] (ladder) at (0, 0) {
            \includegraphics[width=.2\linewidth]{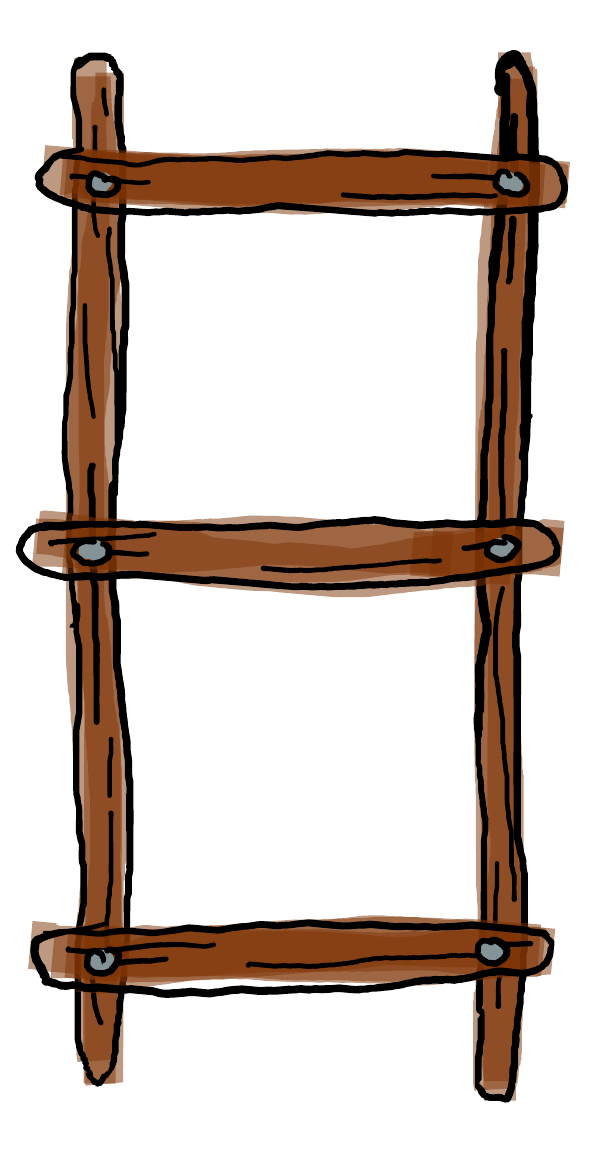}
        };
        \node[inner sep=0, align=center] at ($(ladder) + (0, 3.3)$) {\large \bf Ladder of causation \citep{pearl2018book}};
        
        \node[inner sep=0, align=center] at ($(ladder) + (9.8, 3.3)$) {\large \bf Map of CDL};
        \node[inner sep=0, align=center, text width=4.5cm] at ($(ladder) + (9.8, 2.2)$) {\textcolor{purple}{\it We refine the ladder by including a scale to categorise model parameters and structure categories.}};
        \node[inner sep=0, align=center] at ($(ladder) + (.05, 2.5)$) {\footnotesize counterfactual};
        \coordinate (cf) at ($(ladder) + (2.5, 2.2)$);
        \node[inner sep=0, align=center] at ($(ladder) + (0, .58)$) {interventions};
        \coordinate (iv) at ($(ladder) + (2.5, .2)$);
        \node[inner sep=0, align=center] at ($(ladder) + (0, -1.65)$) {association};
        \coordinate (as) at ($(ladder) + (2.5, -2.1)$);

        \node[example, label=\citep{vaswani2017attention}] (transf) at ($(as) + (-.1, .3)$) {};
        \node[example, label=\citep{williams2006gaussian}] (rf) at ($(as) + (.3, -.2)$) {};
        \node[example, label=\citep{montgomery2021introduction}] (lr) at ($(as) + (-.5, -.1)$) {};

        \node[circle, very thick, draw=purple, inner sep=1.2, fill=white,  label={above:\textcolor{purple}{\it all are categorised the same!}}] (note) at ($(as) + (2, 2)$) {};
        \draw[->, thick, purple] (note) to[out=270, in=0] ($(as) + (.7, .1)$);

        \node (input_knowledge_A) at (9.5,-4.5) {}; 
        \draw[grid] ($(input_knowledge_A) + (0, .5)$) -- ($(input_knowledge_A) + (4, .5)$);
        \node (input_knowledge_U) at ($(input_knowledge_A) + (0, 1)$) {};
        \draw[grid] ($(input_knowledge_U) + (0, .5)$) -- ($(input_knowledge_U) + (4, .5)$);
        \node (input_knowledge_C) at ($(input_knowledge_U) + (0, 1)$) {};
        \node at ($(input_knowledge_U) + (-.3, 0)$) {\rotatebox{90}{structure}};
        \node at ($(input_knowledge_A) + (2, -.8)$) {parameters};
        
        \node (input_param_NO) at (10,-5) {};
        \draw[grid] ($(input_param_NO) + (.5, 0)$) -- ($(input_param_NO) + (.5, 3)$);
        \node (input_param_1) at ($(input_param_NO) + (1, 0)$) {};
        \draw[grid] ($(input_param_1) + (.5, 0)$) -- ($(input_param_1) + (.5, 3)$);
        \node (input_param_2) at ($(input_param_1) + (1, 0)$) {};
        \draw[grid] ($(input_param_2) + (.5, 0)$) -- ($(input_param_2) + (.5, 3)$);
        \node (input_param_M) at ($(input_param_2) + (1, 0)$) {};

        \draw[very thick] ($(input_knowledge_A) - (0, .5)$) -- ($(input_knowledge_C) + (0, .5)$) -- ($(input_knowledge_C) + (4, .5)$) -- ($(input_param_M) + (.5, 0)$) -- cycle;

        \node[example, label=\citep{williams2006gaussian}] at ($(input_knowledge_A) + (1.5, -.2)$) {};
        \node[example, label=\citep{vaswani2017attention}] at ($(input_knowledge_A) + (.5, -.2)$) {};
        \node[example, label=\citep{montgomery2021introduction}] at ($(input_knowledge_A) + (2.5, -.2)$) {};

        \draw[purple, thick, ->] ($(as) + (.4, -.4)$) to[out=-35, in=225] node[midway, text width=2.5cm, fill=white, align=center] {\textcolor{purple}{\it the map refines categorisation}} ($(input_knowledge_A) + (.5, -.8)$);

    \end{tikzpicture}
    \caption{{\bf Refining the ladder of causation.} The ladder (left) depicts different levels of causal knowledge. The higher one climbs, the more knowledge one needs. However, the ladder does not differentiate between (at the extreme) non-parametric models or highly parametric models. Furthermore, we find there are more structures to be taken into account beyond causal structures. Details on the axes are provided in \cref{sec:map}.}
    \label{fig:PCH:ladder}
    \rule{\textwidth}{.5pt}
\end{figure}

The practice of categorising methods according to causal knowledge is not new. In fact, Pearl has introduced the widely adopted ``ladder of causation'' \citep{pearl2018book,bareinboim2022pearl} which is illustrated in \cref{fig:PCH:ladder} (on the left).

{\bf Deep learning.} Topics such as identification and reasoning are much less studied in deep learning \citep{pearl2019seven}. Instead, the focus on deep learning is a much more pragmatic one. Deep learning research is concerned with increasing accuracy \citep{hestness2019beyond,cho2015much,litjens2016deep,krizhevsky2017imagenet,deng2009imagenet,russakovsky2015imagenet,berrevoets2021disentangled}, improving quality of synthetic data \citep{nikolenko2021synthetic,rombach2022high,kobyzev2020normalizing,van2021decaf}, or dealing with an increasing amount of parameters \citep{hunter2012selection,bebis1994feed,mostafa2019parameter}. These advances are completely ignored by Pearl's ladder (as it only focuses on the causal knowledge embedded in the models). This fact is illustrated by including some examples of models that are parameterised very differently (from linear regression \citep{montgomery2021introduction} to transformer networks \citep{vaswani2017attention}.

{\bf Causal deep learning (CDL).} Clearly both causality and deep learning have their own respective merits. One emphasises research on mitigating bias or providing rigorous estimates, whereas the other is much more pragmatic and focuses on empirical results. However, recently, researchers start to pay much more attention on the interaction between each field.

Given the two basic pillars of causal deep learning: (i) causality, and (ii) deep learning, we propose to categorise causal deep learning methods according to these two pillars. While it is clear that each field can aid the other, the focus of each pillar is actually conflicting, e.g., typical work in causality is very good at leveraging structure, but not very good at non-parametric estimation; the reverse is true for deep learning. We propose to measure exactly how each method in this new and emerging field compares to one another by considering how they match up in each of CDL's pillars. To do this, we introduce a map of causal deep learning.

To illustrate its necessity, we have included a rudimentary version of the map of CDL in \cref{fig:PCH:ladder}. While Pearl's ladder is not able to differentiate between transformer networks and linear regression, our map {\it is}. Naturally, we discuss in much more detail how we arrive at our map in \cref{sec:map}. However, \cref{fig:PCH:ladder} shows (on the right) that our map {\it can} indeed differentiate between these wildly different examples.

{\bf A living paper.} We will constantly update this paper to incorporate new developments, accommodating the growth and many changes this field undergoes going forward. However, we believe we have constructed our map to stand the test of time.

\section{The structural and parametric scale}

We propose a map to organise past and future research in CDL. 
Our map helps us to understand and categorise the many differences and similarities between models in CDL, but also to understand how to situate CDL (and these models) in the wider field of machine learning. 
Given the increasing amount of new ideas and proposals in CDL, we believe this to be a necessary exercise.

Causal deep learning refines machine learning by leveraging a different set of assumptions and inductive biases. As these are vital to CDL, we build our map around their classification. Specifically, a these assumptions and biases are defined by two ingredients: (i) the structure linking individual variables, and (ii) the functions that model these links. 

As such, the map of causal deep learning operates under two main axes: (i) {\it the structural scale} concerning the links between variables, described in \cref{sec:scale}; and (ii) {\it the parametric scale} concerning the shape of the functions that model links between variables, described in \cref{sec:param}.

Our two scales form the basis of the map of causal deep learning, which we introduce in detail in \cref{sec:map}.

\begin{figure}[t]
    \centering
    \begin{tikzpicture}[
        dot/.style={circle, fill=black, draw=black,  minimum size=2mm, inner sep=0},
    ]
    
    \node[inner sep=0] (start) at (0, 0) {};
    \node[inner sep=0] (end) at (8, 0) {};
    
    \node[dot, inner sep=0] (blob) at (start) {};
    \node[anchor=south, inner sep=0] at ($(start) - (0, .5)$) {\textcolor{purple}{unknown}};
    \begin{scope}[shift={($(blob) + (0,.5)$)}]
        \node[inner sep=1] (S) at (-.8, 0) {$S$};
        \node[inner sep=1] (C) at (0, 0) {$C$};
        \node[inner sep=1] (D) at (.8, 0) {$D$};
        
        \draw (S) -- (C);
        \draw (C) -- (D);
        \draw (D) to [out=135, in=45] (S);
    \end{scope}

    \node[dot, inner sep=0] (directed) at ($(start)!.5!(end)$) {};
    \node[anchor=south, inner sep=0] at ($(start)!.5!(end) - (0, .565)$) {\textcolor{purple}{plausible}};
    \begin{scope}[shift={($(directed)+(0,.5)$)}]
        \node[inner sep=1] (S) at (-.8, 0) {$S$};
        \node[inner sep=1] (C) at (0, 0) {$C$};
        \node[inner sep=1] (D) at (.8, 0) {$D$};
        
        \draw[->, densely dotted] (C) to [in=45, out=135] (S);
        \draw[->, densely dotted] (C) to [out=45, in=135] (D);
        
        \draw[->] (S) -- (C);
        \draw[->] (C) -- (D);
        
        \draw[->, densely dashed] (D) to [out=225, in=315] (C);
        \draw[->, densely dashed] (C) to [out=225, in=315] (S);
    \end{scope}
    
    \node[dot, inner sep=0] (causal) at (end) {};
    \node[anchor=south, inner sep=0] at ($(end) - (0, .5)$) {\textcolor{purple}{causal}};
    \begin{scope}[shift={($(causal)+(0,.5)$)}]
        \node[inner sep=1] (S) at (-.8, 0) {$S$};
        \node[inner sep=1] (C) at (0, 0) {$C$};
        \node[inner sep=1] (D) at (.8, 0) {$D$};

        \draw[->] (S) -- (C);
        \draw[->] (C) -- (D);
    \end{scope}
    
    \node[blue] at ($(start) + (0, 2.25)$) {{\bf little} structure};
    \node[blue] at ($(end) + (0, 2.25)$) {{\bf lots of} structure};
    
    \draw[very thick] (start) -- (end);

    \draw[thick, blue, latex-latex, rounded corners=3mm] ($(start)+(0, 1.15)$) -- ($(start) + (0, 1.65)$) -- ($(end) + (0, 1.65)$) -- ($(end) + (0,1.15)$);
    
    \end{tikzpicture}
    \caption{{\bf The structural scale.} A first axis on the map of causal deep learning is the structural scale which categorises different structure types. The two extremes are non-informative structures, or a complete causal graph. Note that the structural scale makes no parametric assumptions on these structures (parametric assumptions are the focus of the parametric scale in \cref{fig:param}).}
    \label{fig:scale}
    \rule{\textwidth}{.5pt}
\end{figure}
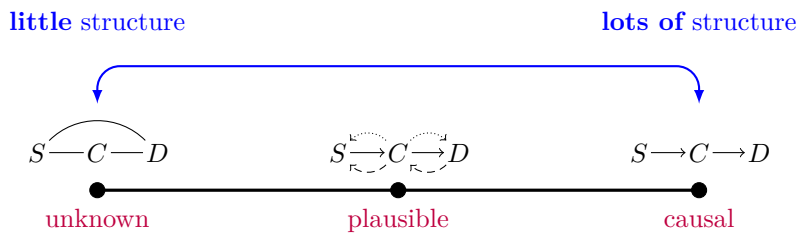

\subsection{The structural scale} \label{sec:scale}
Crucial to causal deep learning is the ability to leverage knowledge about the relationships between variables. 
Depicted as graphs, we can place these relations on a scale from less detailed to more detailed -- this gives the structural scale, which is central to our map of causal deep learning.
%

\cref{fig:scale} depicts the structural scale, which ranges from \textcolor{blue}{{\bf less} structure}, to \textcolor{blue}{{\bf lots of} structure}. The structural scale governs information about the statistical or causal relationship in the environment of interest. Essentially, it quantifies how much we know, expressed in graphical structures. Below, we discuss each level  using an illustrative example of three variables: smoking, denoted as $S$; cancer, denoted as $C$; and death, denoted as $D$.

\textbf{\em Level 1 -- no structure.} At the lowest end of our scale we have structures that exhibit zero knowledge. These structures connect each variable with each other. An example of such a structure is denoted as, 
\begin{equation} \label{eq:blob}
    \begin{tikzpicture}[
        baseline=-1.1mm
    ]
        \node[inner sep=1] (S) at (-.8, 0) {$S$};
        \node[inner sep=1] (C) at (0, 0) {$C$};
        \node[inner sep=1] (D) at (.8, 0) {$D$};
        
        \draw (S) -- (C);
        \draw (C) -- (D);
        \draw (D) to [out=135, in=45] (S);
        
        \node at (4, 0) {\textcolor{purple}{\bf (unknown).}};
    \end{tikzpicture}
\end{equation}
While technically a structure, we have no idea about the potential relationships these variables have with respect to each other. Having a fully connected structure such as in \cref{eq:blob} means that we assume {\it potential} dependence between each variable. Assuming a fully connected structure is not a strict assumption. In our example above, the structure in \cref{eq:blob} basically states that ``smoking, cancer and death'' are possibly related, which is a very general statement indeed.

That is because relatedness is a flexible term. While having an edge may appear to be a strict statement, the edge does not express the {\it level} at which two variables are related. In fact, the {\it absence} of an edge is a much stronger assumption which we discuss next.

\textbf{\em Level 2 -- plausibly causal structures.} Let us assume now that we know a little more: cancer is the reason why smoking is related to death. Statistically, we can express this assumption as $S \independent D | C$. In Level 2, we assume {\it statistical independence}. Independence is not directly the same as assuming some structural information, however, it can be expressed as such.

The easiest thing we can do to reflect $S \independent D | C$ is to explicitly by remove the edge \begin{tikzpicture}[
    baseline=-1.1mm
]
    \node[inner sep=1] (S) at (-.8, 0) {$S$};
    \node[inner sep=1] (D) at (.8, 0) {$D$};

    \draw (D) to [out=135, in=45] (S);
\end{tikzpicture} from \cref{eq:blob} as we have below in \cref{eq:association},
\begin{equation} \label{eq:association}
    \begin{tikzpicture}[
        baseline=-1.1mm
    ]
        \node[inner sep=1] (S) at (-.8, 0) {$S$};
        \node[inner sep=1] (C) at (0, 0) {$C$};
        \node[inner sep=1] (D) at (.8, 0) {$D$};
        
        \draw (S) -- (C) -- (D);
        
    \end{tikzpicture}.
\end{equation}

Crucially, however, there are more structures that reflect $S \independent D | C$. Consider the three structures below,

\begin{minipage}{\textwidth}
\vspace{5pt}
    \begin{minipage}[t]{.32\textwidth}
        \centering
        \begin{tikzpicture}
            \node[inner sep=1] (S) at (-.8, 0) {$S$};
            \node[inner sep=1] (C) at (0, 0) {$C$};
            \node[inner sep=1] (D) at (.8, 0) {$D$};
            
            \draw[->] (S) -- (C);
            \draw[->] (C) -- (D);
        \end{tikzpicture},
    \end{minipage}
    ~
    \begin{minipage}[t]{.32\textwidth}
        \centering
        \begin{tikzpicture}
            \node[inner sep=1] (S) at (-.8, 0) {$S$};
            \node[inner sep=1] (C) at (0, 0) {$C$};
            \node[inner sep=1] (D) at (.8, 0) {$D$};
            
            \draw[<-] (S) -- (C);
            \draw[->] (C) -- (D);
        \end{tikzpicture},
    \end{minipage}
    ~
    \begin{minipage}[t]{.32\textwidth}
        \centering
        \begin{tikzpicture}
            \node[inner sep=1] (S) at (-.8, 0) {$S$};
            \node[inner sep=1] (C) at (0, 0) {$C$};
            \node[inner sep=1] (D) at (.8, 0) {$D$};
            
            \draw[<-] (S) -- (C);
            \draw[<-] (C) -- (D);
        \end{tikzpicture},
    \end{minipage}
\vspace{1pt}
\end{minipage}
which all model the same independencies.

For the sake of our discussion, let us assume $S \independent D | C$ is correct\footnote{If cancer renders smoking and death independent, we imply that cancer is the {\it only} path from smoking to death, which is known to be false \citep{carter2015}. For example, smoking may result in chronic obstructive pulmonary disease (COPD) or ischemic heart disease, which can also lead to premature death. However, for the purpose of illustration, we assume it is true.}. As (in)dependence is a statement of statistical association, we cannot assume that the structures based on it are causal in nature. For example, \begin{tikzpicture}[
    baseline=-1.1mm
]
    \node[inner sep=1] (S) at (-.8, 0) {$S$};
    \node[inner sep=1] (C) at (0, 0) {$C$};
    \node[inner sep=1] (D) at (.8, 0) {$D$};
    
    \draw[<-] (S) -- (C);
    \draw[<-] (C) -- (D);
\end{tikzpicture}, implies that dying impacts whether or not the deceased is a smoker, which cannot be true.

The type of structure of interest in Level 2, are the probabilistic graphical models (PGMs) \citep{koller2009probabilistic}. These include structures such as directed acyclic graphs (DAGs), undirected graphs, or even factor graphs. Depending on the set of independence statements one wishes to assume, one may express them as a certain type of PGM. 

For example, if one wishes to assume $S \independent D$ while also assuming $S \not\independent D | C$, we best use a DAG:
\begin{equation} \label{eq:collider}
    \begin{tikzpicture}
        \node[inner sep=1] (S) at (-.8, 0) {$S$};
        \node[inner sep=1] (C) at (0, 0) {$C$};
        \node[inner sep=1] (D) at (.8, 0) {$D$};
        
        \draw[->] (S) -- (C);
        \draw[<-] (C) -- (D);
    \end{tikzpicture},
\end{equation}
which is often called a ``collider'' structure. This collider structure (and more importantly the (in)dependence statements it implies) can only be modeled with a DAG.

Being able to model independence is a clear step up from Level 1, where we assumed nothing at all. While independence is a statement of statistical association (i.e. it is bidirectional and not causal), they do pose constraints for a later causal model. As we can check (in)dependence in data, a recovered causal model should respect (or even explain) these (in)dependence statements. Hence, the above structures are plausibly causal.

In level 2 we have no consensus on one graph in particular, only about dependence and independence (predominately statistical concepts). For the purpose of illustrating Level 2 on our structural scale in \cref{fig:scale} above, we denote these plausible structures as,
\begin{equation} \label{eq:mec}
\begin{aligned}
    \begin{tikzpicture}[
        baseline=-1.1mm
    ]
        \node[inner sep=1] (S) at (-.8, 0) {$S$};
        \node[inner sep=1] (C) at (0, 0) {$C$};
        \node[inner sep=1] (D) at (.8, 0) {$D$};
        
        \draw[->, densely dotted] (C) to [in=45, out=135] (S);
        \draw[->, densely dotted] (C) to [out=45, in=135] (D);
        
        \draw[->] (S) -- (C);
        \draw[->] (C) -- (D);
        
        \draw[->, densely dashed] (D) to [out=225, in=315] (C);
        \draw[->, densely dashed] (C) to [out=225, in=315] (S);
        
        \node at (4, 0) {\textcolor{purple}{\bf (plausible).}};
    \end{tikzpicture}
    \end{aligned}
\end{equation}
where we combined all possible DAGs that respect, $S \independent D | C$ \citep{sun2006causal}. Applying the rules of d-separation \citep{verma1990}.

While Level 2 includes all types of PGMs, we chose to illustrate Level 2 using DAGs only, as DAGs are the structure of choice when modelling causality \citep{pearl2009causality}. In \cref{eq:mec} all of the structures are DAGs, but only one of them is causal (likely to be the first one).

\textbf{\em Level 3 -- causal structure.} That leaves assuming the exact causal structure as opposed to the many plausible causal structures in Level 2. With \cref{eq:causation} we say that smoking {\it causes} cancer which causes death. Such knowledge translates into the directed structure,
\begin{equation} \label{eq:causation}
    \begin{tikzpicture}[
        baseline=-1.1mm
    ]
        \node[inner sep=1] (S) at (-.8, 0) {$S$};
        \node[inner sep=1] (C) at (0, 0) {$C$};
        \node[inner sep=1] (D) at (.8, 0) {$D$};

        \draw[->] (S) -- (C);
        \draw[->] (C) -- (D);
        
        \node at (4, 0) {\textcolor{purple}{\bf (causal).}};
    \end{tikzpicture}
\end{equation}
Which was indeed one of the plausible DAGs in Level 2 (\cref{eq:mec}).

Comparing \cref{eq:association} with \cref{eq:causation} may lead to believe there is little difference between either end of the structural scale-- {\it nothing could be further from the truth \citep{pearl2018book,bareinboim2022pearl}!} Rather, the step from association to causation is very large and is a central argument in causality. So large in fact that we recognise several intermediary structures.

An important realisation is that each level of information further specifies the environment, i.e. with information, we {\it restrict} possible interpretations and data-distributions. For example, from \cref{eq:blob,eq:association} we expect $S$ to change if $C$ is changed as both variables are associated with each other. However, \cref{eq:causation} tells us this is not necessarily the case: patients do not automatically start smoking when diagnosed with cancer. However, the reverse {\it is} true; when a cancer-patient {\it stops} smoking they have a higher chance of remission \citep{sheikh2021postdiagnosis}.

{\bf ``Rung 1.5''} Pearl's ladder of causation \citep{pearl2018book,bareinboim2022pearl} ranks structures in a similar way as we do, i.e., increasing a model's causal knowledge will yield a higher place upon his ladder. Like Pearl, we have three different levels in our scale. However, they do not correspond one-to-one. 

In particular, we find that our Level 2 is not well represented in Pearl's ladder of causation\footnote{Similarly, his counterfactual rung is not represented in our structural scale.}. Our reasoning is that our Level 2 {\it does} encode some prior knowledge into a model, which is more than encoding no prior knowledge at all. Yet, both would be categorised under Pearls first rung. Of course, we recognise that the models categorised under Level 2 have no embedded {\it causal} knowledge. While Level 2 does not reach the level of a rung 2 model (but Level 3 does), we consider a Level 2 model to correspond with a hypothetical ``rung 1.5''.

\subsection{The parametric scale} \label{sec:param}

\begin{figure}[t]
    \centering
    \begin{tikzpicture}[
        dot/.style={circle, fill=black, draw=black,  minimum size=2mm, inner sep=0},
    ]
    
    \node[dot, label={below:non-param.}] (0moment) at (0, 0) {};
    \begin{scope}[shift={($(0moment) + (0,.5)$)}]
        \node[inner sep=1] at (0, 0) {$f(\mathbf{X}, \mathbf{U}_\mathbf{X})$};
    \end{scope}
    
    \node[dot, label={below:fully known}] (multmoment) at (8, 0) {};
    \begin{scope}[shift={($(multmoment) + (0,.5)$)}]
        \node[inner sep=1] at (0, 0) {$f^*(\mathbf{X}, \mathbf{U}_\mathbf{X})$};
    \end{scope}
    
    \node[dot, label={below:Noise models}] (1moment) at ($(0moment)!.333!(multmoment)$) {};
    \begin{scope}[shift={($(0moment)!.333!(multmoment) + (0,.5)$)}]
        \node[inner sep=1] at (0, 0) {$g(\mathbf{X}) +  \mathbf{U}_\mathbf{X}$};
    \end{scope}
    
    \node[dot, label={below:fully param.}] (2moment) at ($(0moment)!.666!(multmoment)$) {};
    \begin{scope}[shift={($(0moment)!.666!(multmoment) + (0,.5)$)}]
        \node[inner sep=1] at (0, 0) {$g(\mathbf{X})$};
    \end{scope}

    \draw[very thick] (0moment) -- (1moment) -- (2moment) -- (multmoment);

    \node[DarkGreen] at ($(0moment) + (0, 1.75)$) {{\bf no} assumptions};
    \node[DarkGreen] at ($(multmoment) + (0, 1.75)$) {{\bf many} assumptions};
    \draw[thick, DarkGreen, latex-latex, rounded corners=3mm] ($(0moment)+(0, .75)$) -- ($(0moment) + (0, 1.25)$) -- ($(multmoment) + (0, 1.25)$) -- ($(multmoment) + (0,.75)$);

    \end{tikzpicture}
    \caption{{\bf The parametric scale.} A second axis in the map of causal deep learning. The parametric scale logs the type of assumptions made on the factors of the assumed distribution or model. The extremes are no assumptions at all (leaving completely non-parametric factors), or a fully known model. The parametric scale further discerns assumptions on the way noise interacts in the system ($\eps$) and functional shape of the system's factorisation.}
    \label{fig:param}
    \rule{\textwidth}{.5pt}
\end{figure}
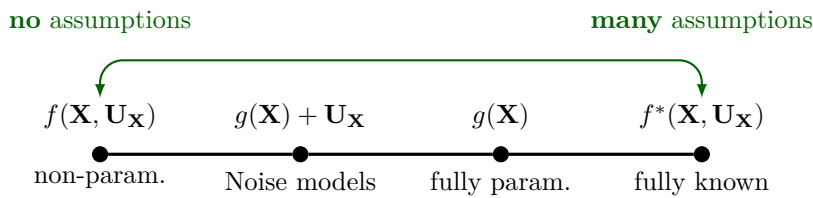

In \cref{sec:scale} we defined structures governing the (statistical or causal) relationships between variables. Now we turn to the parameterisation of these relationships. Intuitively, the structures depict which variables express an {\it affinity} with each other \citep[Chapter~4]{koller2009probabilistic}. In particular, each structure dictates how a distribution can be factorised as a series of functions. These functions are called {\it factors}, which are defined as,
\begin{definition}[factor]
Let $\mathbf{X}$ be a set of random variables. We define a factor $f$ to be a function from $\text{Val}(\mathbf{X}) \to \mathbb{R}$. A factor is nonnegative if all its entries are nonnegative. The set of variables $\mathbf{X}$ is called the scope of the factor and denoted $\text{Scope}[f]$ \citep[Definition~4.1]{koller2009probabilistic}.
\end{definition}
As such, the parametric scale--- which we introduce in this section ---models not {\it if} variables interact, but {\it how} they interact. For example, a statement such as $p \models (S \independent D | C)$ holds iff we factorise the distribution as $p(S, C, D) = f_1(S, C)f_2(C, D)$ (lacking a factor $f_3(S, D)$). As such, a structure becomes parameterised by associating a set of factors with it.

Note that associating factors do not necessarily correspond to associating functions for each edge, at least in the undirected setting (such as \cref{eq:association}). Rather, a factor can be generalised by allowing them over arbitrary subsets of variables. As such, in its most general form, we define a factorisation as,
\begin{equation}\label{eq:factor}
    p(\mathbf{X}) = \frac{1}{Z} \prod_{i\in[I]} f_i(\mathbf{X}^{(i)}),
\end{equation}
where $Z$ is a normalising constant, and one factor $f_i$ takes a set of variables, $\mathbf{X}^{(i)} \subseteq \mathbf{X}$, as arguments \citep{koller2009probabilistic}.

A trivial example of such a factorisation is the decomposition of a joint distribution into conditional distributions: $p(A, B) = f_1(A, B)f_2(B)$ with $f_1(A, B) = p(A | B)$ and $f_2 = p(B)$. With \cref{eq:factor} one can generalise decomposition to accommodate parameterisation of symmetrical relationships as well as directional relationships (which are generally modelled as conditional probability densities).

Notice that in \cref{eq:factor} we iterate over $I$ factors. However, the amount of factors is, at this stage, not yet fully determined. For example, one could have a factorisation which is comprised of the least amount of factors: $p(\mathbf{X}) = f(\mathbf{X})$, or one could have one where we hope to include a factor with only $X^1$ as an argument: $p(\mathbf{X}) = p(X^1)p(X^2|X^1)...p(X^n|X^{n-1}, ..., X^1)$.

Defining a factor, $f_i$ as in \cref{eq:factor}, is non-parametric (we assume nothing about the factors) and agnostic to the structure it models (because there is no implied correspondence between edges and factors). What {\it does} change is the arguments that need to go into a factor. Those {\it are} determined by the accompanying structure\footnote{While not the topic of discussion here, it is this realisation that is often used as an argument in favour of using causal knowledge to increase efficiency by reducing the input required for correct inference (e.g. in \citep{lee2018structural, lee2019structural, kyono2020castle, van2021decaf, hassanpour2019learning}).}.

In our example above, having $A \independent B$ would imply that $p(A | B) = p(A)$, which actively alters the variables that need to go into $f_1$. Recall, that Level 1 of the structural scale is not able to model this explicitly-- we need at least a Level 2 assumption for this.

\textbf{\em Level 1 -- non-parametric.} As with the structural scale, Level 1 of the parametric scale encompasses the least strict assumption: namely, there exists a factorisation such as \cref{eq:factor}. Beyond that, level 1 assumes no specific functional form of its factorisations, $f_i$. 

Of course, as we move up the structural scale, the structures themselves may dictate the purpose of the factorisation. For example, in a directed structure, the set of arguments changes from an arbitrary set $\mathbf{X}^{(i)}$ to the set containing a variable, its parents and noise: $f_X(X, \pa(X), U_X)$. Nevertheless, despite the factor's arguments changing the function remains non-parametric.

\textbf{\em Level 2 -- noise models.} The next level in our parametric scale is to first make an assumption on the composition of the set of random variables, $\mathbf{X}$. In particular, we assume that $\mathbf{X}$ can be separated in noise terms or exogenous variables $\mathbf{U}$ and variables $\mathbf{X}\setminus\mathbf{U}$ where each element in $\mathbf{U}$ corresponds with exactly one element in $\mathbf{X}\setminus\mathbf{U}$ and $|\mathbf{X}| = 2|\mathbf{U}|$.

The above assumption has implications for the factors. Specifically, if a factor takes $X \in \mathbf{X}\setminus\mathbf{U}$ as argument, it will always take $U_X \in \mathbf{U}$ (which is the corresponding noise variable of $X$) as argument also.

Note that the above is {\it only} an assumption on the factor arguments and is per our earlier discussion a {\it graphical} assumption. As such, currently, Level 2 is no different from level 1 in the parametric scale. However, in Level 2, we make an additional assumption which {\it is} a parametric assumption on the factors. For a factor $f_i(\mathbf{X}^{(i)}, \mathbf{U}_{\mathbf{X}^{(i)}})$ we make an assumption on how the noise variables $\mathbf{U}_{\mathbf{X}^{(i)}}$ are incorporated in $f_i$ \citep{zhang2012identifiability,shimizu2011directlingam,hoyer2008nonlinear}, for example, additive noise:
\begin{equation} \label{eq:add}
    f(\mathbf{X}, \mathbf{U}_\mathbf{X}) = g(\mathbf{X}) + \mathbf{U}_\mathbf{X}.
\end{equation}

In short, \cref{eq:add} assumes we can decompose a factor $f$ into the noise, $\mathbf{U}_\mathbf{X}$, and some (non-parametric) function, $g(\mathbf{X})$. While the way in which $\mathbf{U}_\mathbf{X}$ interacts in $f$ is fixed, $g$ is still non-parametric and therefore only level 2 of this scale. As discussed below, we find many papers in causal deep learning making assumptions of this level. 

When performing causal discovery, making an assumption on the noise of a system may aid in causal identification \citep{peters2017elements}. Conceptually, making this type of assumption is another way to exert some outside expert knowledge into the system. In \citet{hoyer2008nonlinear} $g$ remains non-linear (through the use of neural networks) but $\mathbf{U}_\mathbf{X}$ is assumed additively independent (as above) resulting in a clear level 2 assumption.

\textbf{\em Level 3 -- fully parametric.} Naturally, the next level is to make a parametric assumption on $g$ in \cref{eq:add}. These assumptions vary from making linear assumptions, to $n$ times differentiable, to many more types of assumptions on the functional form of $g$ and consequentially $f$. Note that because we make the assumption on $g$ (and not $f$ directly), we automatically assume that $f$ can be decomposed in $g$ and noise, i.e. Level 3 subsumes assumptions from Level 2.

Less parametric assumptions imply a larger model hypothesis class. For example, having linear assumptions allows much less possible factors than a non-linear class of assumptions. In fact, it has been shown that neural networks\footnote{In particular with a ReLU activation.} have a connection to non-parametric regression, leading to high performance models \citep{schmidt2020nonparametric}.

As such, making a parametric assumption is typically done in favour of easier regression, but at the cost of less general model classes.

\textbf{\em Level 4 -- fully known factors.} Rather than making assumptions on the functional class of the factors, some methods assume full access to a known set of factors which we term $f^*(\mathbf{X}, \mathbf{U}_\mathbf{X})$. This is of course a very strict assumption, resulting in the final level of the parametric scale.

Having defined our two axes, we can now compose the complete {\it map of causal deep learning} in \cref{sec:map} which governs both model input as well as model output.

\section{The map of causal deep learning} \label{sec:map}

Consider some data, $\mathcal{D} \coloneqq \{\mathbf{X}_i : \mathbf{X}_i \in \mathcal{X}, i \in [N]\}$, with $\mathcal{X}$ some space such as $\mathbb{R}^d$. Generally, the goal of deep learning is to represent these data using a model, 
\begin{equation} \label{eq:cdl_model}
    m: \mathcal{D} \to \mathcal{H},
\end{equation}
with $m$ the model, and $\mathcal{H}$ a suitably defined space of learned representations which depends on the method to learn $m$.
At a later stage, $\mathcal{H}$ is used to perform a task (such as prediction, or data-generation, or decision making). To perform this task accurately, one has to carefully consider: (i) the type of data ($\mathcal{D}$), and (ii) the type of representation ($\mathcal{H}$) upon which we base the model's output.

The same holds for models in causal deep learning. The difference, however, lays in the types of assumptions and restrictions we pose on (i) the data, and (ii) the representation. Naturally, despite being classified under the umbrella term causal deep learning, there are many differences in how these restrictions manifest. As such, the map of causal deep learning maps (i) the data, and (ii) the learnt representation.

Note that in many cases the representation is the same as the model output. An example of this are causal discovery methods where the aim is to structure the input data as a causal graph. This structure is of course a ``representation''. Specifically, a causal discovery method will transform a dataset into a graph: $m: \mathcal{D} \to \mathcal{G}$. In the latter case, we have that $\mathcal{G} = \mathcal{H}$, and $\mathcal{H}$ and the output are the same, corresponding with \cref{eq:cdl_model}.

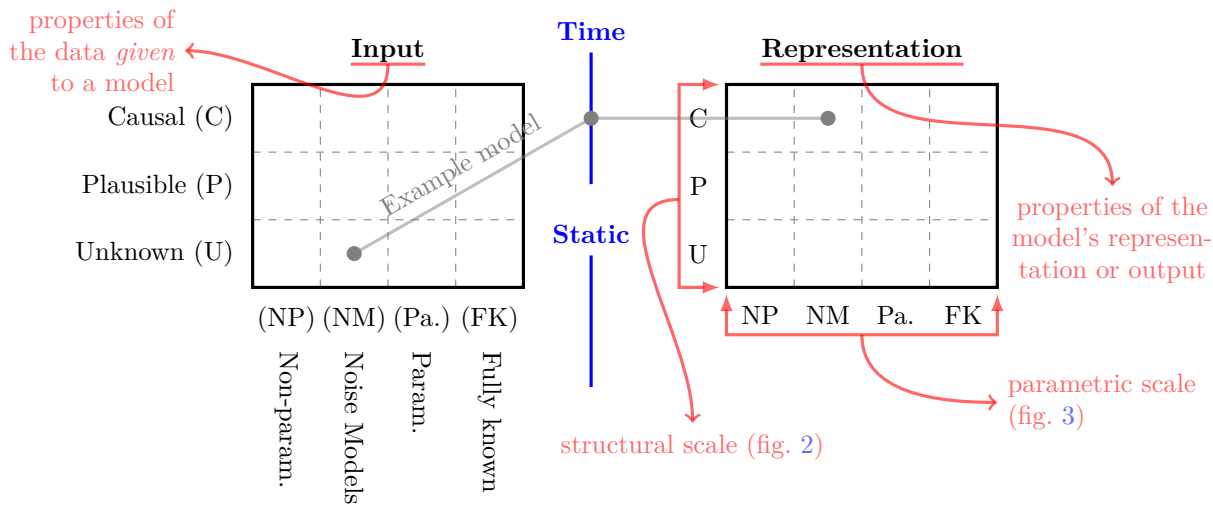
\begin{figure}[t]
    \centering
    \newcommand\Square[1]{+(-#1,-#1) rectangle +(#1,#1)}
    \begin{tikzpicture}[
        grid/.style={gray, dashed},
        annotation/.style={very thick, red, opacity=.6},
        dot/.style={circle, fill=black, line width=0,  minimum size=2mm, inner sep=0},
        route/.style={very thick, draw opacity=.5},
        scale=.9
    ]
        \node[label={left:Unknown (U)}] (input_knowledge_A) at (0,0) {}; 
        \draw[grid] ($(input_knowledge_A) + (0, .5)$) -- ($(input_knowledge_A) + (4, .5)$);
        \node[label={left:Plausible (P)}] (input_knowledge_U) at ($(input_knowledge_A) + (0, 1)$) {};
        \draw[grid] ($(input_knowledge_U) + (0, .5)$) -- ($(input_knowledge_U) + (4, .5)$);
        \node[label={left:Causal (C)}] (input_knowledge_C) at ($(input_knowledge_U) + (0, 1)$) {};
        
        \node[label=below:(NP)] (input_param_NO) at (.5,-.5) {};
        \node[rotate=-90, anchor=west] at (.5, -1.3) {Non-param.};
        \draw[grid] ($(input_param_NO) + (.5, 0)$) -- ($(input_param_NO) + (.5, 3)$);
        \node[label=below:(NM)] (input_param_1) at ($(input_param_NO) + (1, 0)$) {};
        \node[rotate=-90, anchor=west] at ($(input_param_NO) + (1, -.8)$) {Noise Models};
        \draw[grid] ($(input_param_1) + (.5, 0)$) -- ($(input_param_1) + (.5, 3)$);
        \node[label=below:(Pa.)] (input_param_2) at ($(input_param_1) + (1, 0)$) {};
        \node[rotate=-90, anchor=west] at ($(input_param_1) + (1, -.8)$) {Param.};
        \draw[grid] ($(input_param_2) + (.5, 0)$) -- ($(input_param_2) + (.5, 3)$);
        \node[label={below:{(FK)}}] (input_param_M) at ($(input_param_2) + (1, 0)$) {};
        \node[rotate=-90, anchor=west] at ($(input_param_2) + (1, -.8)$) {Fully known};

        \draw[very thick] ($(input_knowledge_A) - (0, .5)$) -- ($(input_knowledge_C) + (0, .5)$) -- ($(input_knowledge_C) + (4, .5)$) -- ($(input_param_M) + (.5, 0)$) -- cycle;
        
        \node (input_title) at ($(input_param_NO)!.5!(input_param_M) + (0, 3.5)$) {\bf Input};

        
        \node[inner sep=0] (time_start) at ($(input_param_M) + (1.5, 3.5)$) {};
        \node[inner sep=0] (time_end) at ($(input_param_M) + (1.5, 1.5)$) {};
        \node[inner sep=0] (static_start) at ($(input_param_M) + (1.5, .5)$) {};
        \node[inner sep=0] (static_end) at ($(input_param_M) + (1.5, -1.5)$) {};
        
        \draw[very thick, blue] (time_start) -- (time_end);
        \draw[very thick, blue] (static_start) -- (static_end);
        
        \node at ($(time_start) + (0, .3)$) {\bf \textcolor{blue}{Time}};
        \node at ($(static_start) + (0, .3)$) {\bf \textcolor{blue}{Static}};

        \node[label={left:U}] (repr_knowledge_A) at ($(input_knowledge_A) + (7, 0)$) {}; 
        \draw[grid] ($(repr_knowledge_A) + (0, .5)$) -- ($(repr_knowledge_A) + (4, .5)$);
        \node[label={left:P}] (repr_knowledge_U) at ($(repr_knowledge_A) + (0, 1)$) {};
        \draw[grid] ($(repr_knowledge_U) + (0, .5)$) -- ($(repr_knowledge_U) + (4, .5)$);
        \node[label={left:C}] (repr_knowledge_C) at ($(repr_knowledge_U) + (0, 1)$) {};
        
        \node[label={below:NP}] (repr_param_NO) at ($(repr_knowledge_A) + (.5, -.5)$) {};
        \draw[grid] ($(repr_param_NO) + (.5, 0)$) -- ($(repr_param_NO) + (.5, 3)$);
        \node[label={below:NM}] (repr_param_1) at ($(repr_param_NO) + (1, 0)$) {};
        \draw[grid] ($(repr_param_1) + (.5, 0)$) -- ($(repr_param_1) + (.5, 3)$);
        \node[label={below:Pa.}] (repr_param_2) at ($(repr_param_1) + (1, 0)$) {};
        \draw[grid] ($(repr_param_2) + (.5, 0)$) -- ($(repr_param_2) + (.5, 3)$);
        \node[label={below:FK}] (repr_param_M) at ($(repr_param_2) + (1, 0)$) {};

        \draw[very thick] ($(repr_knowledge_A) - (0, .5)$) -- ($(repr_knowledge_C) + (0, .5)$) -- ($(repr_knowledge_C) + (4, .5)$) -- ($(repr_param_M) + (.5, 0)$) -- cycle;
        
        \node (repr_title) at ($(repr_param_NO)!.5!(repr_param_M) + (0, 3.5)$) {\bf Representation};

        
        \draw[annotation] ($(input_title) - (.55, .2)$) -- ($(input_title) + (.55, -.2)$);
        \draw[->, annotation] ($(input_title) - (0, .2)$) to[out=270, in=0] ($(input_title) - (3, 0)$) node[annotation, text width=3cm, align=right, anchor=east] {properties of the data {\it given} to a model};

        \draw[annotation] ($(repr_title) - (1.5, .2)$) -- ($(repr_title) + (1.5, -.2)$);
        \draw[->, annotation] ($(repr_title) - (0, .2)$) to[out=270, in=90] ($(repr_title) + (3.7, -2)$) node[annotation, text width=3cm, align=center, anchor=north] {properties of the model's representation or output};
        
        \draw[latex-latex, annotation] ($(repr_knowledge_C) + (-.1, .5)$) -- ($(repr_knowledge_C) + (-.7, .5)$) -- ($(repr_knowledge_A) + (-.7, -.5)$) -- ($(repr_knowledge_A) + (-.1, -.5)$);
        \draw[->, annotation] ($(repr_knowledge_C)!.6!(repr_knowledge_A) + (-.7, 0)$) to[out=180, in=90] ($(repr_param_NO) + (-1, -2)$) node[annotation, text width=4cm, align=center, anchor=north] {structural scale (\cref{fig:scale})};
        
        \draw[latex-latex, annotation] ($(repr_param_NO) - (.5, .1)$) -- ($(repr_param_NO) - (.5, .7)$) -- ($(repr_param_M) + (.5, -.7)$) -- ($(repr_param_M) + (.5, -.1)$);
        \draw[->, annotation] ($(repr_param_NO)!.5!(repr_param_M) + (0,-.7)$) to[out=270, in=180] ($(repr_param_M) + (.5, -1.7)$) node[annotation, text width=3cm, align=left, anchor=west] {parametric scale (\cref{fig:param})};

        \draw[gray, route] ($(input_knowledge_A) + (.5, 0) + 1*(1, 0)$) node[dot, gray] {} -- node[midway, above, rotate=30, gray, opacity=1] {Example model} ($(time_start) - (0, 1)$) node[dot, gray] {} -- ($(repr_knowledge_C) + (.5, 0) + 1*(1, 0)$) node[dot, gray] {};
            
    \end{tikzpicture}
    \caption{{\bf The map of causal deep learning.} The map evaluates each model based on the required input and learnt representations. For each, we categorise using both the structural scale (\cref{fig:scale}) and the parametric scale (\cref{fig:param}). Furthermore, we also discern between models that handle temporal or static data. As an example, we show a fictitious method which assumes no structure but {\it does} make an additive noise assumption; operates in the temporal domain; and provides a truly causal representation under the same additive noise assumption. Our map allows easy examination of a method on many properties at once.}
    \label{fig:map:annotated}
    \rule{\textwidth}{.5pt}

\end{figure}

Using our scales in \cref{fig:scale,fig:param}, we provide some categorisation of both objects, leading to {\it the map of causal deep learning}, illustrated in \cref{fig:map:annotated}. The map of causal deep learning has three components: Input, Time/static, and Representation. Input and Representation are built from the scales we introduced above. The Time/Static component is a simple binary structure which helps us discern methods that operate on temporal data or static data, respectively.

{\bf Input.} The input field (leftmost grid in \cref{fig:map:annotated}) categorises the data in terms of assumptions on the data-generating process (DGP). Here, the structural scale measures how much we already know about the DGP in terms of {\it which} variables are dependent; whereas the parametric scale measures {\it how} these variables are dependent. 

\textbf{\textcolor{blue}{Time/Static}.} We explicitly differentiate between temporal and static models. While evaluated similarly, time may lead to alternative considerations in the causal setting. For example, a typical assumption in the temporal domain is that causes precede effects \citep{granger1969investigating}. While we do not express any opinion on whether or not this makes things harder or easier, we do believe it is important to differentiate in order to properly compare methods. Furthermore, when using our map to search for an appropriate method from a practical point of view, it makes sense to search across static methods for a static problem, and temporal methods for a temporal problem.

{\bf Representation.} A model's representation is evaluated using the same scales as the {\bf Input} of the model. This is a deliberate choice as we wish to document how much information a model adds to existing data.

In terms of representation, the structural scale measures what type of structure a model yields; whereas the parametric scale measures how the model is parameterised. Note that the latter is different from assumptions on the DGP as it concerns the maximum capacity of a particular model.

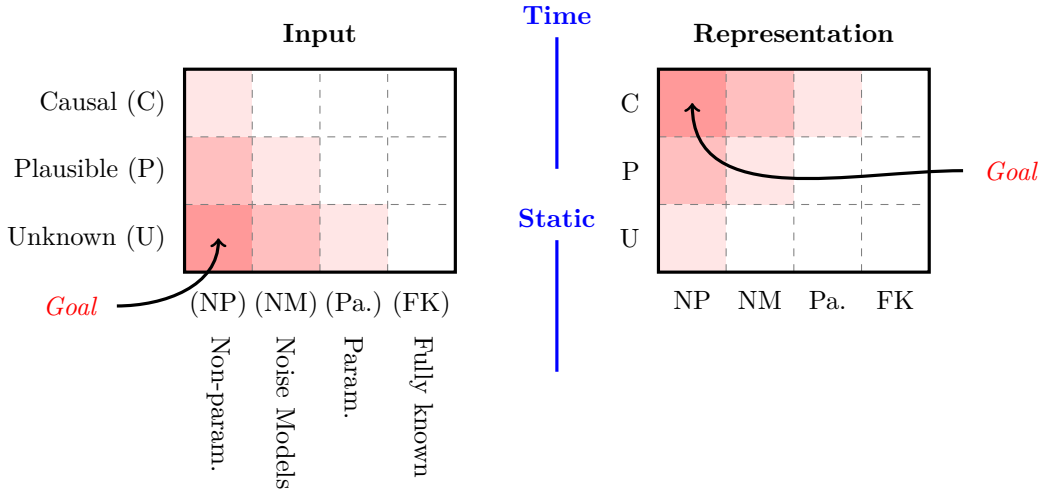
\begin{figure}[t]
    \centering
    \newcommand\Square[1]{+(-#1,-#1) rectangle +(#1,#1)}
    \begin{tikzpicture}[
        grid/.style={gray, dashed},
        scale=.9
    ]
        \node[label={left:Unknown (U)}] (input_knowledge_A) at (0,0) {}; 
        \draw[grid] ($(input_knowledge_A) + (0, .5)$) -- ($(input_knowledge_A) + (4, .5)$);
        \node[label={left:Plausible (P)}] (input_knowledge_U) at ($(input_knowledge_A) + (0, 1)$) {};
        \draw[grid] ($(input_knowledge_U) + (0, .5)$) -- ($(input_knowledge_U) + (4, .5)$);
        \node[label={left:Causal (C)}] (input_knowledge_C) at ($(input_knowledge_U) + (0, 1)$) {};
        
        \node[label=below:(NP)] (input_param_NO) at (.5,-.5) {};
        \node[rotate=-90, anchor=west] at (.5, -1.3) {Non-param.};
        \draw[grid] ($(input_param_NO) + (.5, 0)$) -- ($(input_param_NO) + (.5, 3)$);
        \node[label=below:(NM)] (input_param_1) at ($(input_param_NO) + (1, 0)$) {};
        \node[rotate=-90, anchor=west] at ($(input_param_NO) + (1, -.8)$) {Noise Models};
        \draw[grid] ($(input_param_1) + (.5, 0)$) -- ($(input_param_1) + (.5, 3)$);
        \node[label=below:(Pa.)] (input_param_2) at ($(input_param_1) + (1, 0)$) {};
        \node[rotate=-90, anchor=west] at ($(input_param_1) + (1, -.8)$) {Param.};
        \draw[grid] ($(input_param_2) + (.5, 0)$) -- ($(input_param_2) + (.5, 3)$);
        \node[label={below:{(FK)}}] (input_param_M) at ($(input_param_2) + (1, 0)$) {};
        \node[rotate=-90, anchor=west] at ($(input_param_2) + (1, -.8)$) {Fully known};

        \draw[very thick] ($(input_knowledge_A) - (0, .5)$) -- ($(input_knowledge_C) + (0, .5)$) -- ($(input_knowledge_C) + (4, .5)$) -- ($(input_param_M) + (.5, 0)$) -- cycle;
        
        \node (input_title) at ($(input_param_NO)!.5!(input_param_M) + (0, 3.5)$) {\bf Input};

        
        \node[inner sep=0] (time_start) at ($(input_param_M) + (2, 3.5)$) {};
        \node[inner sep=0] (time_end) at ($(input_param_M) + (2, 1.5)$) {};
        \node[inner sep=0] (static_start) at ($(input_param_M) + (2, .5)$) {};
        \node[inner sep=0] (static_end) at ($(input_param_M) + (2, -1.5)$) {};
        
        \draw[very thick, blue] (time_start) -- (time_end);
        \draw[very thick, blue] (static_start) -- (static_end);
        
        \node at ($(time_start) + (0, .3)$) {\bf \textcolor{blue}{Time}};
        \node at ($(static_start) + (0, .3)$) {\bf \textcolor{blue}{Static}};

        \node[label={left:U}] (repr_knowledge_A) at ($(input_knowledge_A) + (7, 0)$) {}; 
        \draw[grid] ($(repr_knowledge_A) + (0, .5)$) -- ($(repr_knowledge_A) + (4, .5)$);
        \node[label={left:P}] (repr_knowledge_U) at ($(repr_knowledge_A) + (0, 1)$) {};
        \draw[grid] ($(repr_knowledge_U) + (0, .5)$) -- ($(repr_knowledge_U) + (4, .5)$);
        \node[label={left:C}] (repr_knowledge_C) at ($(repr_knowledge_U) + (0, 1)$) {};
        
        \node[label={below:NP}] (repr_param_NO) at ($(repr_knowledge_A) + (.5, -.5)$) {};
        \draw[grid] ($(repr_param_NO) + (.5, 0)$) -- ($(repr_param_NO) + (.5, 3)$);
        \node[label={below:NM}] (repr_param_1) at ($(repr_param_NO) + (1, 0)$) {};
        \draw[grid] ($(repr_param_1) + (.5, 0)$) -- ($(repr_param_1) + (.5, 3)$);
        \node[label={below:Pa.}] (repr_param_2) at ($(repr_param_1) + (1, 0)$) {};
        \draw[grid] ($(repr_param_2) + (.5, 0)$) -- ($(repr_param_2) + (.5, 3)$);
        \node[label={below:FK}] (repr_param_M) at ($(repr_param_2) + (1, 0)$) {};

        \draw[very thick] ($(repr_knowledge_A) - (0, .5)$) -- ($(repr_knowledge_C) + (0, .5)$) -- ($(repr_knowledge_C) + (4, .5)$) -- ($(repr_param_M) + (.5, 0)$) -- cycle;
        
        \node (repr_title) at ($(repr_param_NO)!.5!(repr_param_M) + (0, 3.5)$) {\bf Representation};

        \draw[draw=none, fill=red!40, on layer=back] ($(input_knowledge_A) + (.5, 0)$) \Square{.5};
        
        \draw[draw=none, fill=red!25, on layer=back] ($(input_knowledge_A) + (.5, 0) + 1*(1, 0)$) \Square{.5};
        \draw[draw=none, fill=red!25, on layer=back] ($(input_knowledge_U) + (.5, 0)$) \Square{.5};
        
        \draw[draw=none, fill=red!10, on layer=back] ($(input_knowledge_C) + (.5, 0) + 0*(1, 0)$) \Square{.5};
        \draw[draw=none, fill=red!10, on layer=back] ($(input_knowledge_U) + (.5, 0) + 1*(1, 0)$) \Square{.5};
        \draw[draw=none, fill=red!10, on layer=back] ($(input_knowledge_A) + (.5, 0) + 2*(1, 0)$) \Square{.5};

        \draw[draw=none, fill=red!40, on layer=back] ($(repr_knowledge_C) + (.5, 0)$) \Square{.5};
        
        \draw[draw=none, fill=red!25, on layer=back] ($(repr_knowledge_C) + (.5, 0) + 1*(1, 0)$) \Square{.5};
        \draw[draw=none, fill=red!25, on layer=back] ($(repr_knowledge_U) + (.5, 0)$) \Square{.5};
        
        \draw[draw=none, fill=red!10, on layer=back] ($(repr_knowledge_C) + (.5, 0) + 2*(1, 0)$) \Square{.5};
        \draw[draw=none, fill=red!10, on layer=back] ($(repr_knowledge_U) + (.5, 0) + 1*(1, 0)$) \Square{.5};
        \draw[draw=none, fill=red!10, on layer=back] ($(repr_knowledge_A) + (.5, 0)$) \Square{.5};

        \draw[->,  very thick] ($(input_knowledge_A) + (-1, -1)$)  node[label={left:{\it \textcolor{red}{Goal}}}] {} to[out=0, in=270] ($(input_knowledge_A) + (.5, 0)$);

        \draw[->,  very thick] ($(repr_knowledge_C) + (4.5, -1)$)  node[label={right:{\it \textcolor{red}{Goal}}}] {} to[out=180, in=270] ($(repr_knowledge_C) + (.5, 0)$);

    \end{tikzpicture}
    \caption{{\bf Goals and direction of the field.} We can use the map of causal deep learning to specify the future of our field. Darker shades of red indicate a harder, but more desirable goal. In the case of input, we wish to minimise the required graphical input and made assumptions; with which we hope to achieve a more knowledgeable representation with minimal assumptions.}
    \label{fig:goals}
    \rule{\textwidth}{.5pt}

\end{figure}

\section{Navigating the field}

Using the map of causal deep learning we can: (i) categorise and compare literature, as well as (ii) identify some areas that are not well explored. Using these two functionalities, we can perhaps attribute (i) to practitioners, and (ii) to researchers. A practitioner is faced with a problem they wish to solve. Using the map of CDL, the practitioner can map their available data onto the input field and search for a method in the representation field that would solve their needs. Perhaps guided by the required assumptions the practitioner is willing to make, the set of solutions is more digestible than scanning all the literature in CDL. The researcher may use the map differently. Rather than scanning the potential solutions, a researcher can use the map to scan which solutions are still lacking (or more likely, underrepresented).

{\bf Problems and goals.} Learning a structured representation space, while retaining flexible assumptions, is difficult. Consider \cref{fig:goals} where we annotated learning a causally structured representation without any assumptions from arbitrary data with ``\textcolor{red}{\it Goal}''. This learning setup is the most ambitious setup included in our map. It is also long thought to be impossible \citep{glymour2019review,geiger1990logic,meek2013strong,eberhardt2017introduction}.

However, \cref{fig:goals} also shows the relative complexity of other setups. In principle, the less strict assumptions we make, the harder it becomes to narrow down the structure in our representation space. For some problem setups it may be sufficient to learn a flexible structure, which in turn allows for more flexible assumptions on the input. Similarly, if a problem requires a completely identified causal structure, our map shows that one may have make to some strict assumptions on the input. As such, the map of CDL exposes a certain balance between input assumptions and achievable structure.

\subsection{Comparing methods}

To show how one can use the map to compare methodologies, we take {\it supervised learning} as an example. Of course, CDL spans more than only supervised learning and so does our map.

To remind ourselves, supervised learning is a problem where we wish to map an input, to a label \citep{murphy2012machine}:
\begin{equation*}
    m: \mathcal{D} \to \mathcal{Y},
\end{equation*}
where $\mathcal{D}$ is once again a dataset, but now including labels: $\{(X_i, Y_i) : i \in [N]\}$ with $X_i \in \mathcal{X}$ and $Y_i \in \mathcal{Y}$. The labels $Y_i \in \mathcal{Y}$ can be anything from a real variable ($\mathcal{Y} = \mathbb{R}$), to a binary label ($\mathcal{Y} = \{0, 1\}$).

As we discussed in \cref{sec:map}, deep learning methods first map the data to a representation before it is mapped to the outcome-label \citep{goodfellow2016deep}: $m: \mathcal{D} \to \mathcal{H} \to \mathcal{Y}$. Here, $\mathcal{H}$ corresponds to the representation as in \cref{eq:cdl_model}. As such, our discussion here does not concern $\mathcal{Y}$, but is instead focused on $\mathcal{H}$ and the structure it may respect.

{\bf Why use CDL for supervised learning?} A fair question indeed. Typically, supervised learning models are evaluated only on the accuracy a model achieves on a hold-out test set. Yet, there is no immediate reason why a causal representation would help in this regard. Rather than maximising accuracy on a hold-out test set, one may resort to CDL to also achieve high accuracy on a subset that is not representative of the training data. Examples such as these include: domain adaptation \citep{zhang2021learning,kyono2021exploiting}, transfer learning \citep{rojas2018invariant,magliacane2017causal}, interpretability \citep{wang2020proactive,moraffah2020causal,kim2019learning,xu2020causality}, or general robustness \citep{wang2020visual,kyono2019improving,muller2021learning,buhlmann2020invariance}.

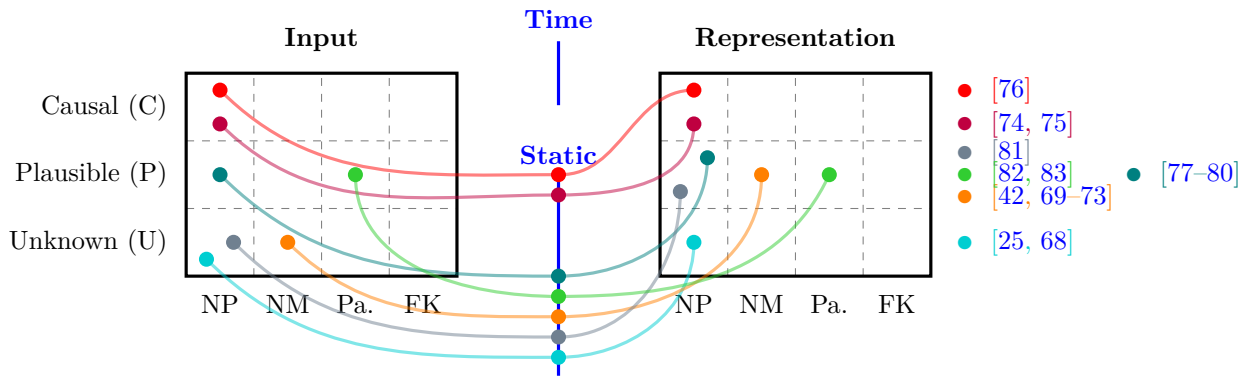
\begin{figure}[t]
    \centering
    \begin{tikzpicture}[
        dot/.style={circle, fill=black, line width=0,  minimum size=2mm, inner sep=0},
        grid/.style={gray, dashed},
        route/.style={very thick, draw opacity=.5},
        scale=.9
    ]
        
        \node[label={left:Unknown (U)}] (input_knowledge_A) at (0,0) {}; 
        \draw[grid] ($(input_knowledge_A) + (0, .5)$) -- ($(input_knowledge_A) + (4, .5)$);
        \node[label={left:Plausible (P)}] (input_knowledge_U) at ($(input_knowledge_A) + (0, 1)$) {};
        \draw[grid] ($(input_knowledge_U) + (0, .5)$) -- ($(input_knowledge_U) + (4, .5)$);
        \node[label={left:Causal (C)}] (input_knowledge_C) at ($(input_knowledge_U) + (0, 1)$) {};
        
        \node[label={below:NP}] (input_param_NO) at (.5,-.5) {};
        \draw[grid] ($(input_param_NO) + (.5, 0)$) -- ($(input_param_NO) + (.5, 3)$);
        \node[label={below:NM}] (input_param_1) at ($(input_param_NO) + (1, 0)$) {};
        \draw[grid] ($(input_param_1) + (.5, 0)$) -- ($(input_param_1) + (.5, 3)$);
        \node[label={below:Pa.}] (input_param_2) at ($(input_param_1) + (1, 0)$) {};
        \draw[grid] ($(input_param_2) + (.5, 0)$) -- ($(input_param_2) + (.5, 3)$);
        \node[label={below:FK}] (input_param_M) at ($(input_param_2) + (1, 0)$) {};

        \draw[very thick] ($(input_knowledge_A) - (0, .5)$) -- ($(input_knowledge_C) + (0, .5)$) -- ($(input_knowledge_C) + (4, .5)$) -- ($(input_param_M) + (.5, 0)$) -- cycle;
        
        \node at ($(input_param_NO)!.5!(input_param_M) + (0, 3.5)$) {\bf Input};

        
        \node[inner sep=0] (time_start) at ($(input_param_M) + (2, 3.5)$) {};
        \node[inner sep=0] (time_end) at ($(input_param_M) + (2, 2.5)$) {};
        \node[inner sep=0] (static_start) at ($(input_param_M) + (2, 1.5)$) {};
        \node[inner sep=0] (static_end) at ($(input_param_M) + (2, -1.5)$) {};
        
        \draw[very thick, blue] (time_start) -- (time_end);
        \draw[very thick, blue] (static_start) -- (static_end);
        
        \node at ($(time_start) + (0, .3)$) {\bf \textcolor{blue}{Time}};
        \node at ($(static_start) + (0, .3)$) {\bf \textcolor{blue}{Static}};

        
        \node (repr_knowledge_A) at ($(input_knowledge_A) + (7, 0)$) {}; 
        \draw[grid] ($(repr_knowledge_A) + (0, .5)$) -- ($(repr_knowledge_A) + (4, .5)$);
        \node (repr_knowledge_U) at ($(repr_knowledge_A) + (0, 1)$) {};
        \draw[grid] ($(repr_knowledge_U) + (0, .5)$) -- ($(repr_knowledge_U) + (4, .5)$);
        \node (repr_knowledge_C) at ($(repr_knowledge_U) + (0, 1)$) {};
        
        \node[label={below:NP}] (repr_param_NO) at ($(repr_knowledge_A) + (.5, -.5)$) {};
        \draw[grid] ($(repr_param_NO) + (.5, 0)$) -- ($(repr_param_NO) + (.5, 3)$);
        \node[label={below:NM}] (repr_param_1) at ($(repr_param_NO) + (1, 0)$) {};
        \draw[grid] ($(repr_param_1) + (.5, 0)$) -- ($(repr_param_1) + (.5, 3)$);
        \node[label={below:Pa.}] (repr_param_2) at ($(repr_param_1) + (1, 0)$) {};
        \draw[grid] ($(repr_param_2) + (.5, 0)$) -- ($(repr_param_2) + (.5, 3)$);
        \node[label={below:FK}] (repr_param_M) at ($(repr_param_2) + (1, 0)$) {};

        \draw[very thick] ($(repr_knowledge_A) - (0, .5)$) -- ($(repr_knowledge_C) + (0, .5)$) -- ($(repr_knowledge_C) + (4, .5)$) -- ($(repr_param_M) + (.5, 0)$) -- cycle;
        
        \node at ($(repr_param_NO)!.5!(repr_param_M) + (0, 3.5)$) {\bf Representation};
        

        \draw[DarkTurquoise, route] ($(input_knowledge_A) + (.3, -.25) + 0*(1, 0)$)  node[dot, DarkTurquoise] {} to[out=315, in=180] ($(static_start)!.9!(static_end)$) node[dot, DarkTurquoise] {} to[out=0, in=270] ($(repr_knowledge_A) + (.5, 0) + 0*(1, 0)$) node[dot, DarkTurquoise] {};
        \node[label={[text=DarkTurquoise]right:\citep{deng2009imagenet,farago1993strong}}] at ($(repr_knowledge_A) + (4.5, 0)$) {\textcolor{DarkTurquoise}{\dott}};
        
        \draw[orange, route] ($(input_knowledge_A) + (.5, 0) + 1*(1, 0)$) node[dot, orange] {} to[out=315, in=180] ($(static_start)!.7!(static_end)$) node[dot, orange] {} to[out=0, in=270] ($(repr_knowledge_U) + (.5, 0) + 1*(1, 0)$) node[dot, orange] {};
        \node[label={[text=orange]right:\citep{kyono2020castle,kyono2021miracle,yao2021path,hollmann2022tabpfn,fatemi2021slaps,morales2021vicause}}] at ($(repr_knowledge_U) + (4.5, -.33)$) {\textcolor{orange}{\dott}};
        
        \draw[purple, route] ($(input_knowledge_C) + (.5, -.25) + 0*(1, 0)$) node[dot, purple] {} to[out=315, in=180] ($(static_start)!.1!(static_end)$) node[dot, purple] {} to[out=0, in=270] ($(repr_knowledge_C) + (.5, -.25)$) node[dot, purple] {};
        \node[label={[text=purple]right:\citep{russo2022causal,teshima2021incorporating}}] at ($(repr_knowledge_C) + (4.5, -.25)$) {\textcolor{purple}{\dott}};
        
        \draw[red, route] ($(input_knowledge_C) + (.5, .25) + 0*(1, 0)$) node[dot, red] {} to[out=315, in=180] (static_start) node[dot, red] {} to[out=0, in=180] ($(repr_knowledge_C) + (.5, .25)$) node[dot, red] {};
        \node[label={[text=red]right: \citep{kancheti22a}}] at ($(repr_knowledge_C) + (4.5, .25)$) {\textcolor{red}{\dott}};

        \draw[Teal, route] ($(input_knowledge_U) + (.5, 0) + 0*(1, 0)$) node[dot, Teal] {} to[out=315, in=180] ($(static_start)!.5!(static_end)$) node[dot, Teal] {} to[out=0, in=270] ($(repr_knowledge_U) + (.7, .25) + 0*(1, 0)$) node[dot, Teal] {};
        \node[label={[text=Teal]right:\citep{kohavi1996scaling,jiang2008novel,jiang2012improving,jiang2016deep}}] at ($(repr_knowledge_U) + (4.5, 0) + (2.5, 0)$) {\textcolor{Teal}{\dott}};

        \draw[SlateGray, route] ($(input_knowledge_A) + (.7, 0) + 0*(1, 0)$)  node[dot, SlateGray] {} to [out=315, in=180] ($(static_start)!.8!(static_end)$) node[dot, SlateGray] {} to[out=0, in=270] ($(repr_knowledge_U) + (.5, 0) + 0*(1, 0) + (-.2, -.25)$) node[dot, SlateGray] {};
        \node[label={[text=SlateGray]right:\citep{NEURIPS2020_e05c7ba4}}] at ($(repr_knowledge_U) + (4.5, .33)$) {\textcolor{SlateGray}{\dott}};
        
        
        \draw[LimeGreen, route] ($(input_knowledge_U) + (.5, 0) + 2*(1, 0)$) node[dot, LimeGreen] {} to [out=270, in=180] ($(static_start)!.6!(static_end)$) node[dot, LimeGreen] {} to[out=0, in=245]  ($(repr_knowledge_U) + (.5, 0) + 2*(1, 0)$) node[dot, LimeGreen] {};
        \node[label={[text=LimeGreen]right:\citep{duda1973pattern,langley1992analysis}}] at ($(repr_knowledge_U) + (4.5, 0)$) {\textcolor{LimeGreen}{\dott}};
        

   \end{tikzpicture}
   
   \caption{{\bf Navigating (static) supervised learning.} Here we focus only on methods that share the task of mapping data to a label in a supervised way. We discern between methods based on what input they expect (or assume) and what type of representation they learn {\it before} mapping to a label. Our map presents a straightforward way to categorise methods in causal deep learning. Doing this is useful for practitioners (to identify a suitable method), and to researchers (to identify a potential research gap). Here too: Non-parametric (NP), Noise models (NM), Parametric (Pa.), and Fully known (FK).}
   \label{fig:supervised_learning}
   \rule{\textwidth}{.5pt}
\end{figure}

\subsection{Developing methods}

Let us discuss our map using a non-exhaustive list of some of the key tasks the map of CDL may help a researcher.

{\bf Identification.} A big part in a researcher's workflow is to identify gaps in the literature. For researchers active in supervised learning, our map of CDL can be a useful tool to do exactly this. For example, from \cref{fig:supervised_learning} we learn that the models that take a fully known function as a parametric assumption are not well represented-- they are not represented at all! Of course, this makes sense. If one assumes a fully known model, before commencing the learning task, there is no point learning it anymore. 

Looking further, we find that mapping additive noise models with unknown structure, to additive noise model {\it with} structure actually {\it are} well represented. The reason for this is that recent contributions in differentiable structure learning are a great candidate to regularise models used for other tasks than structure learning. As such, a researcher would learn that there is indeed much competition in this field.

{\bf Navigation and related works.} This brings us to a second task we can employ our map for: building a body of related works. The map will make it easier for researchers to find and learn about related works. Essentially enabling a platform for researchers to share their work with other interested researches. We explain in our discussion in \cref{sec:discussion} how the map is a living document and how constant updating will help researchers and practitioners in all of the tasks described in this section.

Furthermore, the map allows researchers to think critically about the work they present. For example, in \cref{fig:goals} we clearly state that the ``ultimate goal'' to map data to a non-parametric causal structure without making any assumptions is simply impossible. Not only will this help researchers to discuss their proposal with a sense of realism, the same can be said about practitioners (and reviewers) alike.

{\bf Data.} Beyond the above, the map also provides guidance for model evaluation. In particular, when evaluating a model, it is important to use data that actually matches the models input assumptions. Similarly for the benchmarks the proposal is compared against. 

For example, if a method assumes a causal graph as input, there should be some assurance that the presented graph is indeed causal. One way to do this is by also provided some empirical evidence that this is the case, such as running clinical trials or assuming interventional data.

\subsection{Using methods}

{\bf Navigate.} While the researcher's perspective is mostly based around identifying methods that {\it don't} exist, the practitioner's is the opposite. When faced with a certain problem, the practitioner may use the map to identify which methods assume the input the practitioner has available, while also solving the problem they wish to solve. 

Finding a suitable method can be extremely challenging for practitioners as there are new methods proposed almost daily. This latter point is in fact one of the prime reasons why we propose the map in the first place. 

{\bf Validation and education.} On the other hand, it may be that the practitioner wishes to solve a problem that is not yet solved before (stumbling upon a gap in the literature). Or, more strikingly, they are faced with a problem that is simply impossible to solve.

Especially the latter could be an interesting use of our map: education. We strongly believe that the map of CDL can form a bridge between researches and practitioners. Educating which problems {\it can} be solved (which is communicated from research to practice), and which problems {\it have to be} solved (which is in turn communicated from practice to research). 

As we have observed in \cref{fig:supervised_learning}, we find that there is a heavy focus on learning structure from nothing in the additive noise setting. But, perhaps this is not interesting from a practical point of view as it could be that practitioners are well aware of some conditional independence, or some parametric shape of their environment. Having clear communication between both parties may avoid researchers  to waste valuable resources and time to solve problems that need not be solved.

\section{Discussion} \label{sec:discussion}

We hope that our map organises and identifies the problems, methods, challenges, and future directions of causal deep learning. Doing so may help researchers and practitioners alike. Not only do we believe that the map enables clear communication between both, we thing that using the map will make the tasks associated with both researchers and practitioners much easier. Naturally, the use of our map heavily depends on a sense of maintained relevance. Not only in supervised learning (for which we completed a small exercise in \cref{fig:supervised_learning}), but also fields such as reinforcement learning, generative models, reasoning, etc. As such, we consider this paper to be a {\it living paper}.

{\bf A living paper.} A living paper should be understood as a paper that receives frequent updates. Not only by adding in newly proposed methods, but also by changing the structure of the map to fit the needs of the field of causal deep learning. 
We believe strongly that the only way this intense field can be accurately tracked is by constant monitoring and refining our map. As such, we will provide, in the future, an online forum where we can track, discuss, and propose changes to any part of our paper. This includes the map itself, related work, and our discussion.

\subsubsection*{Acknowledgements}

We wish to already acknowledge Kun Zhang (CMU), Trent Kyono (Meta), and Ruibo Tu (KTH) for discussing this work with us. We truly value their input, as well as future input from community members.

\bibliography{main}
\bibliographystyle{unsrtnat}


\end{document}